\pdfoutput=1
\documentclass[11pt]{article}

\usepackage[]{acl}

\usepackage{times}
\usepackage{latexsym}

\usepackage[T1]{fontenc}
\usepackage[utf8]{inputenc}

\usepackage{microtype}

\usepackage{inconsolata}

\usepackage{amsmath}
\usepackage{amsthm}
\usepackage{bm}
\usepackage{bbm}
\usepackage{mathtools}
\usepackage{booktabs}
\usepackage[ruled, vlined, linesnumbered]{algorithm2e}
\usepackage{subcaption}
\usepackage[switch]{lineno}
\usepackage{xcolor} 
\usepackage{multirow}
\usepackage{mathrsfs} 
\usepackage{balance}
\usepackage{multicol}
\usepackage{makecell}
\usepackage{amssymb}

\usepackage{floatrow}
\newfloatcommand{capbtabbox}{table}[][\FBwidth]

\newcommand{\huimin}[1]{\textcolor{black}{#1}}
\newcommand{\ours}{{Fed-MP}\xspace}

\title{Open-Vocabulary Federated Learning with Multimodal Prototyping}

\author{Huimin Zeng \quad Zhenrui Yue \quad Dong Wang \\
Unversity of Illinois at Urbana-Champaign \\
\{\texttt{huiminz3, zhenrui3, dwang24\}@illinois.edu}}

\begin{document}
\maketitle
\begin{abstract}
% Recently, exploiting vision-language models (e.g., CLIP) for federated learning (FL) has gained attention for their capacity of addressing data heterogeneity in FL. However, existing methods are not tailored for open-vocabulary queries. 

% could not address
% Numerous methods that adapt CLIP for FL applications utilize prompt learning, counting on its unreliable generalization on downstream tasks. More importantly, such learned prompts usually fail to generalize to unseen novel categories, and no proper solution has been proposed. 

{Existing federated learning (FL) studies usually assume the training label space and test label space are identical. However, in real-world applications, this assumption is too ideal to be true. A new user could come up with queries that involve data from unseen classes, and such open-vocabulary queries would directly defect such FL systems.} Therefore, in this work, we explicitly focus on the under-explored open-vocabulary challenge in FL. That is, for a new user, the global server shall understand her/his query that involves arbitrary unknown classes. To address this problem, we leverage the pre-trained vision-language models (VLMs). In particular, we present a novel adaptation framework tailored for VLMs in the context of FL, named as \textbf{Fed}erated \textbf{M}ultimodal \textbf{P}rototyping (\textbf{\ours}). \ours adaptively aggregates {the local model weights} based on light-weight client residuals, and makes predictions based on a novel multimodal prototyping mechanism. \ours exploits the knowledge learned from the seen classes, and robustifies the adapted VLM to unseen categories. Our empirical evaluation on various datasets validates the effectiveness of \ours.
\end{abstract}

\section{Introduction}
\label{sec:intro}
Federated learning (FL) emerges as a new machine learning (ML) paradigm that trains ML models from decentralized data sources \cite{mcmahan2017communication}. {The decentralized nature of FL makes it a promising solution for privacy-sensitive applications across numerous domains (e.g., natural language processing \cite{liu2021federated}, multimodal learning \cite{che2023multimodal}, visual recognition \cite{liu2020fedvision})}. In FL, there exists a central server storing a global model, and a set of clients. The clients will collaboratively train the global model without sharing their private data.  While numerous FL studies have been proposed, the elusive open-vocabulary challenge is largely under-explored. 

\begin{figure}[t]
\centering
\includegraphics[trim=6.4cm 3.6cm 6.4cm 2.4cm, clip, width=0.9\textwidth]{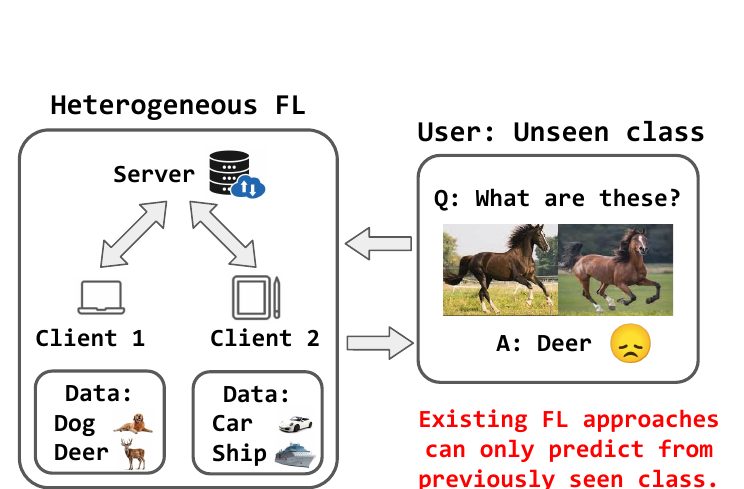}
\caption{A non open-vocabulary FL model could only return a prediction from the seen classes for an open-vocabulary query.}
\label{fig:example}
\vspace{-0.2cm}
\end{figure}

Traditional FL studies (e.g., domain-generalized federated learning) usually assume that the label space of training data and test data is identical. Based on this assumption, the proposed FL methods are not open-vocabulary by design. However, in real-world applications, \textbf{new users} might send queries that involve novel classes, e.g., identifying an object in a photo. If the category of this object is never seen in the training data, then traditional FL systems simply fail and {can only predict from previously seen classes} as shown in Figure~\ref{fig:example}. 

{Indeed, in centralized ML, there exist methods to predict unseen classes \cite{shu2018unseen,he2022safe,changpinyo2017predicting}. However, they usually require a huge amount of the training data and could not tackle new addition of unseen classes over time \cite{kuchibhotla2022unseen}. More importantly, the unique challenge of data heterogeneity in FL makes centralized methods inapplicable to train FL models \cite{jiang2022harmofl,xu2022closing,zhang2023federated}.} {The data heterogeneity in FL is the heterogeneity in client data distributions. For instance, in Figure~\ref{fig:example}, there are only images of dog and deer in  client 1, and client 2 only has images of car and ship. Such non-i.i.d. data across clients is heterogeneous data.} Therefore, in this work, we explicitly focus on the open-vocabulary challenge in FL: how can we build an FL framework that is open-vocabulary?
% {by exploring semantic representations within VLM?}
% the power of language?

On the other hand, exploiting the pre-trained vision-language models (VLMs) (e.g., CLIP) for FL has recently gained increased attention {for their strong generalization ability \cite{lu2023fedclip}}. {With CLIP, the community} could address data heterogeneity, personalization and generalization in FL \cite{lu2023fedclip,yang2023efficient,guo2023pfedprompt}. Technically, to adapt CLIP for specific FL applications, existing methods mainly adopt prompt learning. Prompt learning optimizes a set of learnable soft prompt vectors, and prepends them to input embeddings \cite{lu2023fedclip,yang2023efficient,guo2023pfedprompt}. As such, domain-specific knowledge is integrated into the features extracted by CLIP, leading to improved performance on downstream tasks. Unfortunately, these learned prompts usually suffer from generalizing well to novel unseen classes during test, and yet, no proper solution has been developed. 

% For the open-vocabulary queries in FL applications, one could only count on the unreliable generalization of these learned prompts.

Therefore, in this work, we focus on addressing the elusive open-vocabulary challenge in FL. To the best of our knowledge, we are the first to propose a CLIP-based FL framework that is explicitly tailored for the open-vocabulary setting. To achieve {open-vocabulary FL}, we propose a federated finetuning framework tailored for VLMs: \textbf{Fed}erated \textbf{M}ultimodal \textbf{P}rototyping or \textbf{Fed-MP}. Intuitively, \ours has two design objectives: 1) low communication overhead between the server and clients in FL: given the large size of CLIP, \ours must be light-weight and affordable in terms of model training in an FL application; 2) open-vocabulary: the global model shall understand the queries {that involve arbitrary unseen classes.}

To this end, \ours consists of two modules. Firstly, \ours adaptively aggregates the local model weights based on the similarity between new queries and perturbed client prompt representations. {These prompt representations are perturbed by a set of learnable parameters, which is defined as client residuals. Client residuals protect clients' class information by perturbing the text representations. In addition, with client residuals, locally learned visual concepts are integrated into the perturbed prompt representations as well.} This similarity-based design is realistic and practical in terms of real-world applications: a user comes to use the FL system, and she/he sends a set of queries to the server. In return, the server should adaptively obtain an aggregated model that is aligned with the interest of the user. 
% Client residuals are learnable perturbations that measure the semantic closeness between the user and the clients without sacrificing privacy.
Secondly, we design a multimodal prototyping mechanism to make predictions for the open-vocabulary queries. The multimodal prototypes include text prototypes and visual prototypes. The text prototypes are the original encoded text prompts {in the new queries}. As for the visual prototypes, they are normalized visual features extracted by CLIP image encoder with pseudo labeling. During inference, \ours predicts for a query image based on its weighted distance to text prototypes and visual prototypes. Both modules are designed to exploit the knowledge learned from the seen classes during training. Under \ours, the adapted CLIP model generalizes well to test images from unseen classes, achieving open-vocabulary federated learning.

We summarize the contributions of our paper as follows\footnote{We adopt publicly available datasets and release the code at \url{https://github.com/huiminzeng/Fed-MP.git}.}:
\begin{enumerate}

\item To the best of our knowledge, \ours is the first {VLM-based FL framework that explicitly addresses the open-vocabulary challenge in FL applications}. 

\item Technically, to build \ours, we present a novel adaptive aggregation protocol and a novel multimodal prototyping mechanism. 

% Moreover, \ours is parameter-efficient, and affordable for FL applications.

\item Extensive experimental results on 6 image classification datasets suggest that \ours can effectively improve model performance on test data from unseen categories, outperforming the state-of-the-art baselines.
\end{enumerate}
\section{Related Work}
\label{sec:rela}

\subsection{Federated Learning with Domain Generalization}
Domain generalization (DG) in FL aims to improve model's generalization on the unknown test clients or the unknown global data with domain shifts. Due to privacy concerns (no data exchange) and data heterogeneity, existing centralized DG methods become inapplicable and infeasible in FL \cite{jiang2022harmofl,zhang2023federated,xu2022closing,sun2023feature}. Therefore, a few studies start to investigate DG in FL. For instance, \citet{jiang2022harmofl} propose to establish a harmonized feature space on the frequency domain and aggregate local models with flat optima, so that both local shift and global shift could be rectified. 
% In \citep{xu2022closing}, the domain generalization is achieved by selecting the most similar personalized local model for the unseen domain. 
In comparison, for generalization, \citet{zhang2023federated} introduce a variance reduction regularizer to encourage fairness of the generalization gap among the clients. Finally, in \citep{sun2023feature}, feature distribution matching is proposed to learn domain-invariant client features, so that the model generalizes to unseen clients. However, the above methods all assume that the label space of training data and test data is identical: all tested categories have to be seen during training despite domain shifts. In other words, these methods are not open-vocabulary, and could not handle queries with unseen classes. 

\subsection{Federated Learning with Vision-Language Models}
Recently, integrating vision-language models (e.g., CLIP) into FL has gained increased attention for their strong generalization ability. 
% In particular, existing methods that combines CLIP and FL mainly relies on parameter-efficient finetuning techniques to finetune CLIP to the specific FL applications. 
For instance, \citet{guo2023pfedprompt,guo2023promptfl} focus on learning soft textual prompts to personalize CLIP on client data by extending \cite{zhou2022conditional} into the federated setting, whereas \citet{li2023visual} leverage visual prompts to achieve the same goal. In addition to prompt learning, \citet{lu2023fedclip,chen2023feddat,qiu2023text} fine-tune CLIP with light-weight neural networks (i.e., adapters) to adapt CLIP to FL applications. However, the above methods are not deliberately designed for open-vocabulary settings. Even though the method presented in \cite{qiu2023text} was tested with open-vocabulary queries, its performance purely counts on the unreliable generalization of the learned adapter. In comparison, in this work, we explicitly focus on addressing the open-vocabulary challenge in FL, and present the first FL framework that is tailored for open-vocabulary queries.
\section{Preliminaries}
\label{sec:pre}
\subsection{Federated Learning}
\label{sec:fl}
Assume there are $K$ clients in an FL application. For all clients, each data point is characterized by an input feature ${x} \sim \mathcal{X}$ and a label $y \sim \mathcal{Y}$. On client $k$, its local dataset $\mathcal{D}^{(k)}$ is denoted as $\mathcal{D}^{(k)} = \{({x}_1^{(k)}, y_1^{(k)}), ...|({x}_i^{(k)}, y_i^{(k)}) \sim p^{(k)}\}$, where $p^{(k)}$ represents the local data distribution on client $k$. \textbf{For simplicity, if not specified, we use the notations without the client index $k$ to represent an arbitrary client.}

To find the optimal global model $f_\theta^{*}$ in an FL application, \citet{mcmahan2017communication} propose Federated Averaging (FedAvg). Under FedAvg, at each round, each local client firstly receives a copy of the global model $f_{\theta}$ from the central server and trains the model with its own data. This leads to different local models $(f^{(1)}_{\theta}, f^{(2)}_{\theta}, ...,f^{(K)}_{\theta})$. Then, clients send the trained model weights to the central server. Finally, on the central server, the global model will be updated using a weighted-average of the received model weights based on the size of each local dataset. 

Note that, the local data distributions on different clients could be non-i.i.d. and have exclusive label spaces. 
% For instance, in Figure~\ref{fig:example}, there are only images of dog and deer client 1, and client 2 only has images of car and ship.
% if client 1 is a dog person, then there might be only image data of dogs on clients. Meanwhile, assume client 2 is a cat person, then client 2 might only have cat images. 
% Such heterogeneity in client data distributions is defined as data heterogeneity across clients in FL.
More importantly, in a real-world application, a new user of the FL system might send queries that involve objects from unseen categories. {For instance, in Figure~\ref{fig:example}, the training classes are dog, deer, car, and ship, whereas the test query is an image of horse.}
% That is, in such open-vocabulary queries, the test data might be from categories that have never been observed in the training data of all clients. 

\subsection{CLIP: Contrastive Language-Image Pre-training}
CLIP is a language-grounded image classifier. It predicts which images are paired with which texts. 
Formally, we use $f_{I}$ to denote the CLIP image encoder, and $f_{T}$ for the CLIP text encoder. The inference process and training process of CLIP are:

\begin{itemize}
    \item \textbf{Inference}: For a query image $x$ and $|\mathcal{Y}|$ classes, we firstly craft a set of \textbf{candidate prompts} that contain class information (e.g., \{\texttt{a photo of [class 1]}, \texttt{a photo of [class 2]...}\}). Then, CLIP encodes $x$ into a visual representation $z$, and encodes the candidate prompts into text representations $\{t_{{candidate}_{1}}, t_{{candidate}_{2}}, ..., t_{{candidate}_{{|\mathcal{Y}|}}}\}$. After computing cosine similarity between the $z$ and candidate prompt representations, CLIP selects the prompt with the highest cosine similarity as the final prediction:
    \begin{equation}
    \label{eq:clip_inf}
        \begin{split}
            % & \hat{y} = \arg\max_{c}\frac{\mathrm{exp}\big{(}\mathrm{cos}(z,t_{{candidate}_c})\big{)}}{\sum_{c'}\mathrm{exp}\big{(}\mathrm{cos}(z,t_{candidate_{c'}})\big{)}}, \\
            & \hat{y} = \arg\max_{c}\frac{\mathrm{exp}\big{(}\mathrm{cos}(z,t_{{candidate}_c})/\tau\big{)}}{\sum_{c'}\mathrm{exp}\big{(}\mathrm{cos}(z,t_{candidate_{c'}})/\tau\big{)}}, \\
            & \text{where} \; z = f_{I}(x), \\
            &  t_{{candidate}_c} = f_{T}(\texttt{a photo of [class c]}), \\
            &  c\in\{1,2,...,|\mathcal{Y}|\}.
        \end{split}
    \end{equation}
    \item \textbf{Training}: For a training set $\mathcal{D}$, we construct a \textbf{ground truth prompt} $t_{gt_i}$ for each image $x_i$. For $x_i$, its ground truth prompt contains textual description of its class label $y_i$. Then, the CLIP contrastive loss \cite{radford2021learning} is computed over all visual representations $z_i$s and text representations $t_{{gt}_i}$s:
    \begin{equation}    
    \begin{split}
    \mathcal{L}_{CLIP} & = \frac{1}{|\mathcal{D}|}\sum_{i=1}^{|\mathcal{D}|}-\log\frac{e^{{z}_i \cdot t_{gt_i}}}{\sum_{j=1}^{|\mathcal{D}|}e^{{z}_i\cdot t_{gt_j}}} \\
    & + \frac{1}{|\mathcal{D}|}\sum_{i=1}^{|\mathcal{D}|}-\log\frac{e^{{z}_i \cdot t_{gt_i}}}{\sum_{j=1}^{|\mathcal{D}|}e^{{z}_j \cdot t_{gt_i}}}.
    \end{split}
    \end{equation}
\end{itemize}

\section{Algorithm}
\label{sec:alg}

\begin{figure*}[!bht]
\centering
\begin{subfigure}{\textwidth}
\centering
\includegraphics[trim=0.1cm 6.6cm 0.1cm 6.0cm, clip, width=0.9\textwidth]{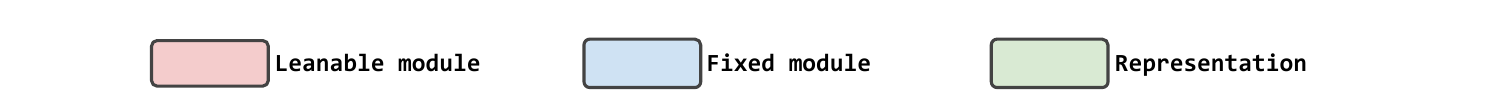}
\end{subfigure}
\begin{subfigure}{\textwidth}
\includegraphics[trim=0.2cm 4.2cm 0.4cm 4.2cm, clip, width=\textwidth]{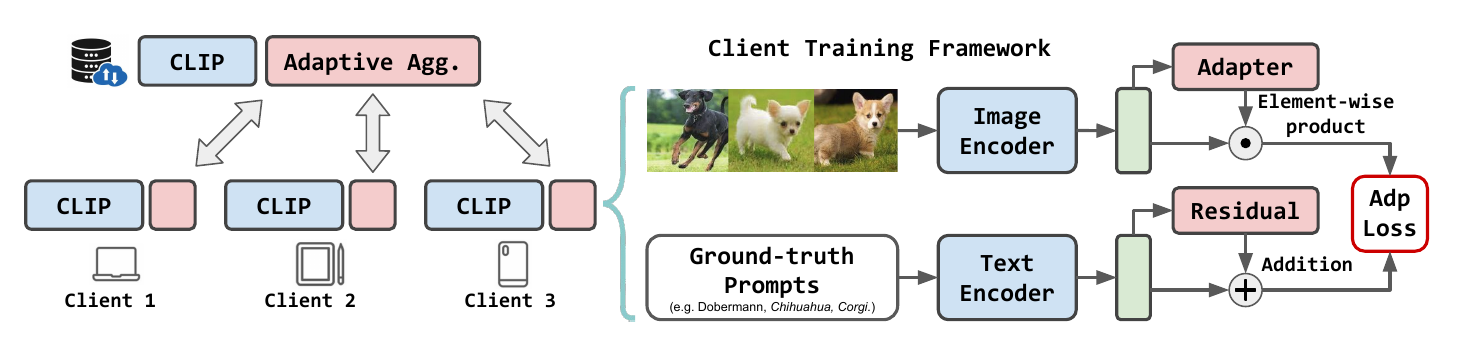}
\end{subfigure}
\caption{The training and aggregation process of \ours. On clients, the adapters and residuals are trained using local data. In adaptive aggregation, only the adapter weights are aggregated.}
\label{fig:overview}
\end{figure*}

% To address open-vocabulary challenges in FL, we propose \ours. The overall framework of \ours is shown in Figure~\ref{fig:overview}. Section~\ref{sec:peft_adaptation} depicts how CLIP could be adapted into FL applications using general multimodal parameter-efficient techniques. In Section~\ref{sec:aggregation}, we demonstrate the first key component of our method, i.e., adaptive aggregation with client residuals. Finally, Section~\ref{sec:prototyping} presents the second key contribution of this work, multimodal prototyping.

\subsection{Parameter-Efficient Adaptation}
\label{sec:peft_adaptation}
Given existing parameter-efficient finetuning (PEFT) methods, any of them could be used by \ours to adapt the CLIP model in FL. In our implementation, we choose to add a small two-layer fully connected network for the visual modality as in \cite{lu2023fedclip}. Formally, we define the adapter as $f_{A}$. As shown in Figure~\ref{fig:overview}, for an input image $x$, $f_{A}$ takes its visual representation as input, i.e., $f_{A}(z), z=f_{I}(x)$. $f_{A}$ returns a vector of normalized importance scores with the same dimensionality of $z$. Finally, the adapted visual representation $z'$ is computed by multiplying $f_{A}(z)$ with $z$ element-wisely:
\begin{equation}
    z' = f_A(z) \odot z, \;\; \mathrm{where} \; z = f_I(x).
\end{equation}
Note that during training, the weights of visual adapter are sent to the global server for aggregation instead of the entire CLIP model.
% , which {reduces the communication overheads}.

% \begin{figure}[!bht]
% \centering
% \includegraphics[trim=7.2cm 5.4cm 7.2cm 4.4cm, clip, width=\textwidth]{figures/adapter.pdf}
% \caption{Adapter}
% \label{fig:adapter}
% % \vspace{-0.2cm}
% \end{figure}

\subsection{Client Residuals}
\label{sec:aggregation}
In an open-vocabulary setting, a new user will send test queries that involve unseen data categories. Thus, to fully exploit the learned knowledge from client data, it is critical to consider the semantic closeness between the clients and the new user when performing model aggregation. Intuitively, the importance weights of local clients should be increased if they are semantically closer to the new user. For instance, client 1 only contains images and prompts of 'Doberman', and client 2 only has images and prompts of 'Tabby cat'. Assume a test query contains an image of a dog, and the candidates prompts are 'a photo of German shepherd' and 'a photo of Welsh Corgi'. In this example, the test class names 'German shepherd' and 'Welsh Corgi' are unseen during training. However, it is intuitive that client 1 is semantically closer to the test query than client 2. The reason is that the prompts of client 1 and the test prompts are all related to dog. Therefore, when aggregating the global model, the importance weight of client 1 should be higher than client 2.

However, existing studies mainly use FedAvg without considering such semantic closeness, and therefore, are not adaptive to open-vocabulary queries. Moreover, directly comparing client class names and the test classes causes privacy leakage: it requires the clients to share class information with the server. Therefore, inspired by \cite{yu2023task}, we proposed to add a set of learnable perturbations to perturb the encoded text prompts for all clients. Such design protects class information on clients. {More importantly, these perturbations will interact with images during training. As such, they provide aligned semantic information from both texts and images.}

% measure the residuals between the clients and the new user: how many residuals are needed to be added to this client, so that this client is semantically identical to the new user in the text feature space.

Formally, we define such perturbations as client residuals. The client residuals on a specific client are a set of learnable perturbations $\Delta = \{\delta_1, \delta_2, ...,\delta_{|\mathcal{Y}|}\}$. {Each $\delta_c \in \Delta$ corresponds to a specific class $c$}, and has the same dimensionality of a prompt representation. When computing the prompt presentations with residuals, CLIP will element-wisely add them to the prompt representations of corresponding classes. For instance, for the \textbf{ground truth prompt} of sample $(x_i, y_i)$, its prompt representation with residual is computed as
\begin{equation}
\begin{split}
    & t^{'}_{{gt}_{i}} = t_{{gt}_{i}} + \alpha \delta_{y_i},
\end{split}
\end{equation} 
where $\delta_{y_i}$ is the perturbation for class $y_i$ (Figure~\ref{fig:overview}), and $\alpha$ is a non-negative scaling factor.

With both trainable adapter and client residuals, the adaptation loss of CLIP on the training set $\mathcal{D} = \{(x_i, y_i)\}$ is computed as follows: 
\begin{equation}
\label{eq:adap_loss}
\begin{split}
    \mathcal{L}_{adp}(f_{A}, \delta) & = \frac{1}{|\mathcal{D}|}\sum_{i=1}^{|\mathcal{D}|}-\log\frac{e^{{z}^{'}_i \cdot t^{'}_{gt_i}}}{\sum_{j=1}^{|\mathcal{D}|}e^{{z}^{'}_i\cdot t^{'}_{gt_j}}} \\
    & + \frac{1}{|\mathcal{D}|}\sum_{i=1}^{|\mathcal{D}|}-\log\frac{e^{{z}^{'}_i \cdot t^{'}_{gt_i}}}{\sum_{j=1}^{|\mathcal{D}|}e^{{z}^{'}_j \cdot t^{'}_{gt_i}}}.
\end{split}
\end{equation}
In Equation ~\ref{eq:adap_loss}, $z^{'}$ represents the adapted visual representation. $t^{'}_{gt}$ is the perturbed text presentation.

After training, the client residuals are added to the encoded candidate prompts, according to the class names. This process returns a set of perturbed representations of candidate prompts:
\begin{equation}
    \mathcal{T}' = \{t^{'}_{{candidate}_{1}}, t^{'}_{{candidate}_{2}}, ..., t^{'}_{{candidate}_{|\mathcal{Y}|}}\},
\end{equation}
where $t^{'}_{{candidate}_{c}} = t_{{candidate}_{c}} + \delta_c$. The client will send $\mathcal{T}'$ to the central server along with the updated adapter. This process will not lead to privacy leakage, as the class names and the training data are not shared with the server. 

% \begin{figure}[!htb]
% \centering
% \includegraphics[trim=7.2cm 5.4cm 6.4cm 4.4cm, clip, width=\textwidth]{figures/residual.pdf}
% \caption{Residual}
% \label{fig:residual}
% % \vspace{-0.2cm}
% \end{figure}

\subsection{Adaptive Model Aggregation with Client Residuals}
After receiving $f^{(1)}_{A}$, $f^{(2)}_{A}$, ..., $f^{(K)}_{A}$ and $\mathcal{T}'^{(1)}$, $\mathcal{T}'^{(2)}$, ..., $\mathcal{T}'^{(K)}$, the central server will then aggregate the adapter weights based on the queries from the new user. The aggregation protocol is based on the similarity between the queries of the new user and the perturbed prompt representations of different clients, namely $\mathcal{T}'^{(1)}$, $\mathcal{T}'^{(2)}$, ..., $\mathcal{T}'^{(K)}$. 

In particular, assume the new user has a set of unlabeled test images $\mathcal{D}_{test}$ and a set of candidate prompts. 
% We further assume the candidate prompts cover all the classes of the entire test label space $\mathcal{Y}_{test}$. 
Note that the test label space $\mathcal{Y}_{test}$ and the client label space $\mathcal{Y}^{(k)}$ is mutually exclusive: $\mathcal{Y}_{test} \cap \mathcal{Y}^{(k)} = \varnothing, k=1,...,K$.

The first step of adaptive aggregation is to encode the test candidate prompts using the CLIP text encoder. This returns a set of prompt representations that correspond the test classes:
\begin{equation}
\begin{split}
    & \mathcal{T}_{test} = \{t_{{test}_1}, t_{{test}_2}, ..., t_{{test}_{|\mathcal{Y}_{test}|}}\}, \\
    & \mathrm{where} \;\;  t_{test_c} = f_{T}(\texttt{a photo of [test class c]}).
\end{split}
\end{equation}
{For instance, in Figure~\ref{fig:example}, test prompts could be \texttt{"a photo of [horse]"} and \texttt{"a photo of [cat]"}, where both \texttt{[horse]} and \texttt{[cat]} are classes never seen during training.}

Next, the server measures the semantic closeness between the new user and all clients. Specifically, it computes the expected similarity between $\mathcal{T}_{test}$ and $\mathcal{T}'^{(1)}$, $\mathcal{T}'^{(2)}$, ..., $\mathcal{T}'^{(K)}$, respectively. For instance, we define the expected similarity between the new user and client $k$ as $\xi_{k}$. It is computed via:
\begin{equation}
\label{eq:simi}
\begin{split}
    & \xi_{k} = \mathbb{E}_{t_{test} \sim \mathcal{T}_{test}, t^{'}_{candidate} \sim \mathcal{T}^{'(k)}} \big[ \mathrm{cos}(t_{test},t'_{candidate}) \big] \\
    & = \frac{1}{|\mathcal{Y}_{test}||\mathcal{Y}^{(k)}|}\sum_{l=1}^{|\mathcal{Y}_{test}|}\sum_{m=1}^{|\mathcal{Y}^{(k)}|}\mathrm{cos}(t_{test_l}, t^{'(k)}_{candidcate_m}).
\end{split}
\end{equation}
Note that  Equation~\ref{eq:simi} computes the averaged cosine similarity between any two encoded prompts, one from the new user and one from client $k$. Moreover, Equation~\ref{eq:simi} does not cause privacy leakage as elaborated in Section~\ref{sec:aggregation}.

After computing the expected similarity for all clients, the server aggregates the adapter weights:
\begin{equation}
\label{eq:fed_avg}
\begin{split}
    & \theta^*_{A} = \frac{1}{\sum_{k} e^{\xi_k}}\sum_{k=1}^K e^{\xi_k} \cdot 
    {\theta}^{(k)}_A.
\end{split}
\end{equation}
In Equation~\ref{eq:fed_avg}, $\theta^*_{A}$ is the aggregated adapter weights. $\theta^{(k)}_{A}$ represents the adapter weights uploaded by client $k$. Compared to FedAvg, Equation~\ref{eq:fed_avg} takes the semantic closeness of the new user and the clients into account. {The rationale behind this design is that semantically closer clients have learned more useful visual concepts related to the open-vocabulary queries, whereas other clients may only learned irrelevant concepts. As such, useful visual concepts should be highlighted and integrated to the adapted CLIP by up-weighting corresponding adpater weights. }

% Moreover, this design leverages the power of language and is tailored for the vision-language models: the CLIP text encoder is capable of extracting semantic meaningful representations from the text inputs.

\subsection{Multimodal Prototyping}
\label{sec:prototyping}

Recall that during inference, for a query image, CLIP will compare the cosine similarity between its visual representation and the representations of candidate prompts (Equation~\ref{eq:clip_inf}). In this context, these prompt representations are by default text prototypes for the test classes. This is because the predictions are produced by measuring the distance (cosine similarity) between the text prototypes and the representation of the input image. Thus, the representations of candidate prompts are defined as the textual prototypes $\{p_1, p_2, ..., p_{|\mathcal{Y}_{test}|}\}$:
\begin{equation}
    \{p_1, p_2, ..., p_{|\mathcal{Y}_{test}|}\}, \; \mathrm{where }\; p_i = t_{test_i}.
\end{equation}
However, the global model has never seen textual prototypes of unseen classes. This leads to poor generalization. 

Therefore, based on the aggregated global model, we further propose to develop a new set of visual prototypes. In particular, inspired by \cite{iwasawa2021test}, for each test class, we define a visual prototype set. Formally, for test class $c$, its visual prototype set is defined as $\mathcal{Q}_{c}$. 

{If the new user send an extensive amount of queries, the global server may need to process them in mini-batches. In this case, we introduce a time stamp $n$ to denote the temporal order of the test process. Meanwhile, the update process follows the same temporal order.} At $n=0$, $Q$s are initialized as empty sets. Then, for a test sample $x$ at time step $n$, the visual prototypes are updated as follows:
\begin{equation}
\label{eq:prototype}
    \mathcal{Q}^{n+1}_{\hat{y}}= 
\begin{cases}
    \mathcal{Q}^{n}_{\hat{y}} \cap \{\frac{z'}{||z'||}\},  & \text{if } \mathcal{H}(x)\leq \epsilon \\
    \mathcal{Q}^{n}_{\hat{y}},              & \text{otherwise}
\end{cases}
\end{equation}
where $z'$ is the adapted representation of $x$. $\hat{y}$ is the pseudo prediction calculated by the adapted CLIP: $\hat{y} = \arg\max_{c}\frac{\mathrm{exp}\big{(}\mathrm{cos}(z',t_{{test}_c})/\tau\big{)}}{\sum_{c'}\mathrm{exp}\big{(}\mathrm{cos}(z',t_{test{c'}})/\tau\big{)}}$. $\mathcal{H}(x)$ is the entropy of the predictive probabilities, evaluating the quality of the prediction: $\mathcal{H}(x) = \sum_{c=1}^{|\mathcal{Y}_{test}|}-P(\hat{y}=c|x)\mathrm{log}P(\hat{y}=c|x)$ as in \cite{iwasawa2021test}. $\epsilon$ is confidence threshold. 

According to Equation~\ref{eq:prototype}, only one prototype set (class $\hat{y}$) would be updated based on the pseudo prediction. Moreover, in our implementation, we implemented Equation~\ref{eq:prototype} in an efficient way, so that there is no need to save all the visual representations (Appendix A). 

Eventually, with the visual prototypes, \ours computes the prediction for the next $x$ based on its distance towards the centroids of the multimodal prototypes. Specifically, under multimodal prototyping, CLIP makes the prediction for $x$ by selecting the closest multimodal prototypes:

\begin{equation}
\label{eq:prototype_inference}
    \begin{split}
        & \hat{y} = \arg\max_{c} \big[ \mathrm{cos}(z', p_c) + \mathrm{cos}(z',\bar{q}_c) \big],
    \end{split}
\end{equation}
where $p_c$ is the textual prototype of \texttt{[test class c]} and $\bar{q}_c$ is the centroid of visual prototypes of \texttt{[test class c]}:
\begin{equation}
\label{eq:centriods}
    \bar{q}_c = \frac{1}{|Q_c|}\sum_{q \in Q_c} q.
\end{equation}

\begin{figure}[t]
\centering
\begin{subfigure}{\textwidth}
\includegraphics[trim=4.8cm 7.2cm 4.8cm 6.2cm, clip, width=\textwidth]{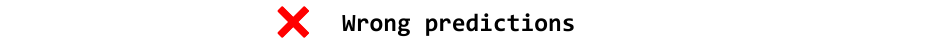}
\end{subfigure}
\begin{subfigure}{0.48\textwidth}
\includegraphics[trim=2.3cm 0.6cm 1.4cm 0cm, clip, width=\textwidth]{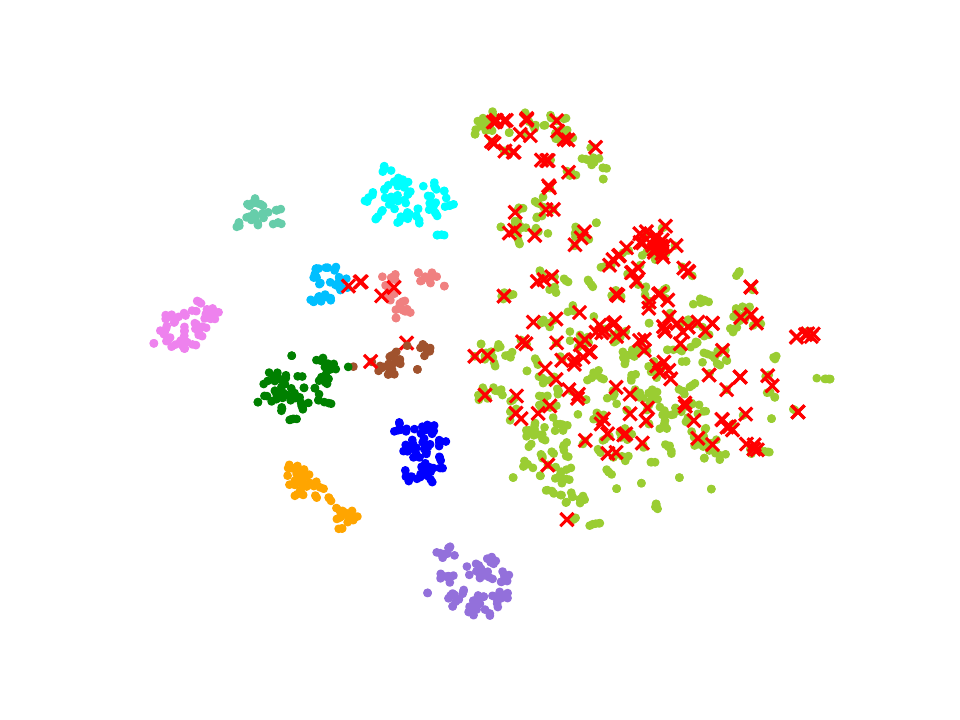}
\caption{w/o Prototyping}
\end{subfigure}
\hfill
\begin{subfigure}{0.48\textwidth}
\includegraphics[trim=2.3cm 0.6cm 1.4cm 0cm, clip, width=\textwidth]{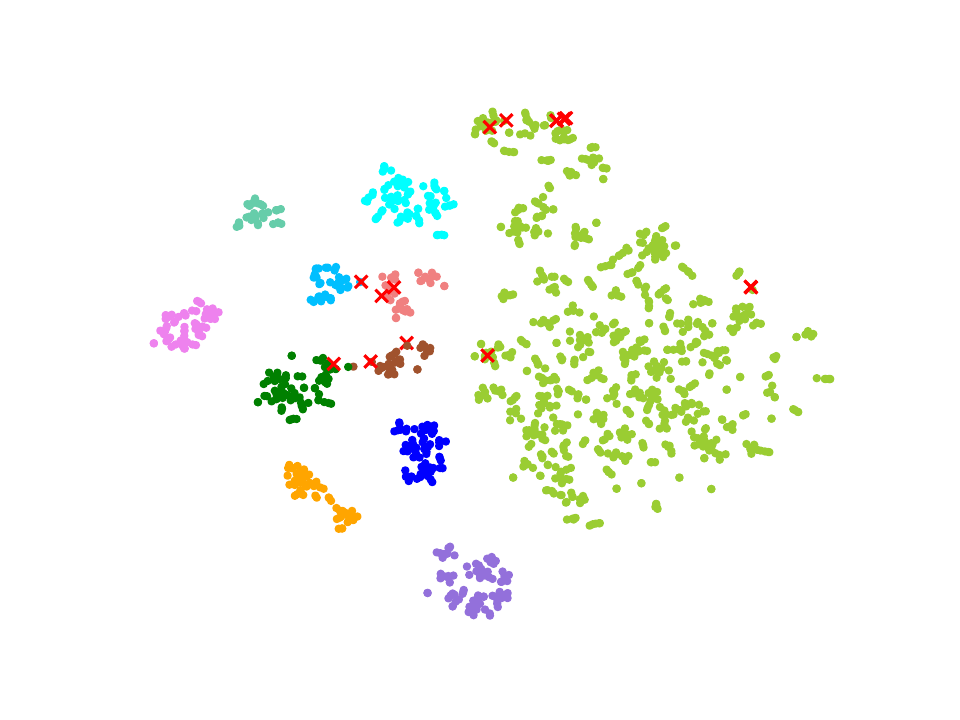}
\caption{with Prototyping}
\end{subfigure}
\caption{T-SNE visualization on test classes from Caltech101.}
\label{fig:tsne}
\end{figure}

The rationale behind multimodal prototyping is: if a test sample obtains a high-quality prediction, then it could serve as a template for other test samples. Moreover, under \ours, the adapted visual representations are semantic-aware, because the global model aggregation is based on the semantic closeness between the clients (training classes) and new user (test classes). Therefore, in addition to textual prototypes, the visual prototypes could also contribute to the model generalization on test data from unseen classes. For instance, in Figure~\ref{fig:tsne} (a), there are many errors for the green class if only textual prototypes are used. In contrast, after performing multimodal prototyping, many wrong predictions are corrected (Figure~\ref{fig:tsne} (b)). The overall framework is summarized in Appendix B.

\section{Experiments}
\label{sec:exp}
We evaluate the proposed \ours mainly on open-vocabulary image classification, {which is one of the prevailing applications for VLMs.} In addition, we also provide an ablation study to understand the function of the modules within \ours. Finally, we conduct robustness studies to evaluate the robustness of \ours in regards to the number of training samples per class. 

% \usepackage{booktabs}
% \usepackage{multirow}
\begin{table*}[!bht]
\small
\resizebox{\textwidth}{!}{
\begin{tabular}{@{}lccccccc@{}}
\toprule
\textbf{Dataset}                     & \textbf{Metrics}                       & FedAvg (NN) & FedKA (NN) & PromptFL & FedTPG & FedCLIP & \ours (ours) \\ 
\midrule
\multirow{3}{*}{Caltech101} & $\mathcal{A}$ $\uparrow$      & $0.5090_{\pm 0.0627}$ & $0.5652_{\pm 0.0526}$ & $\underline{0.9920_{\pm 0.0015}}$ & $0.9909_{\pm 0.0037}$ & $0.9185_{\pm 0.0285}$ & $\textbf{0.9936}_{\pm 0.0010}$      \\
                            & $\Phi_{P}$    $\uparrow$      & $0.6172_{\pm 0.0064}$ & $0.6542_{\pm 0.0472}$ & $0.9799_{\pm 0.0044}$ & $\underline{0.9806_{\pm 0.0043}}$ & $0.8746_{\pm 0.0253}$ & $\textbf{0.9848}_{\pm 0.0030}$      \\
                            & $\Phi_{R}$    $\uparrow$      & $0.6613_{\pm 0.0053}$ & $0.6962_{\pm 0.0477}$ & $\underline{0.9785_{\pm 0.0044}}$ & $0.9721_{\pm 0.0148}$ & $0.9740_{\pm 0.0050}$ & $\textbf{0.9908}_{\pm 0.0014}$      \\
                            & $\Phi_{F1}$   $\uparrow$      & $0.6071_{\pm 0.0047}$ & $0.6472_{\pm 0.0522}$ & $\underline{0.9784_{\pm 0.0047}}$ & $0.9741_{\pm 0.0122}$ & $0.9106_{\pm 0.0213}$ & $\textbf{0.9876}_{\pm 0.0020}$      \\ 
\midrule
\multirow{3}{*}{UCF101}     & $\mathcal{A}$ $\uparrow$      & $0.6491_{\pm 0.0869}$ & $0.6465_{\pm 0.0312}$ & $0.8582_{\pm 0.0093}$ & $0.8473_{\pm 0.0424}$ & $\underline{0.8855_{\pm 0.0178}}$ & $\textbf{0.9127}_{\pm 0.0225}$      \\
                            & $\Phi_{P}$    $\uparrow$      & $0.6622_{\pm 0.0989}$ & $0.6823_{\pm 0.0596}$ & $0.8231_{\pm 0.0038}$ & $0.8168_{\pm 0.0715}$ & $\underline{0.8841_{\pm 0.0258}}$ & $\textbf{0.9212}_{\pm 0.0238}$      \\
                            & $\Phi_{R}$    $\uparrow$      & $0.6491_{\pm 0.0869}$ & $0.6564_{\pm 0.0312}$ & $0.8502_{\pm 0.0093}$ & $0.8473_{\pm 0.0424}$ & $\underline{0.8855_{\pm 0.0178}}$ & $\textbf{0.9127}_{\pm 0.0255}$      \\
                            & $\Phi_{F1}$   $\uparrow$      & $0.6318_{\pm 0.0921}$ & $0.6404_{\pm 0.0385}$ & $0.8318_{\pm 0.0093}$ & $0.8185_{\pm 0.0576}$ & $\underline{0.8760_{\pm 0.0229}}$ & $\textbf{0.9086}_{\pm 0.0298}$      \\ 
\midrule
\multirow{3}{*}{Food101}    & $\mathcal{A}$ $\uparrow$      & $0.5521_{\pm 0.0055}$ & $0.5474_{\pm 0.0046}$ & $0.9240_{\pm 0.0203}$ & $0.9257_{\pm 0.0359}$ & $\underline{0.9719_{\pm 0.0008}}$ & $\textbf{0.9828}_{\pm 0.0005}$      \\
                            & $\Phi_{P}$    $\uparrow$      & $0.5888_{\pm 0.0048}$ & $0.5876_{\pm 0.0038}$ & $0.9438_{\pm 0.0104}$ & $0.9430_{\pm 0.0229}$ & $\underline{0.9731_{\pm 0.0007}}$ & $\textbf{0.9829}_{\pm 0.0005}$      \\
                            & $\Phi_{R}$    $\uparrow$      & $0.5521_{\pm 0.0055}$ & $0.5474_{\pm 0.0046}$ & $0.9240_{\pm 0.0203}$ & $0.9257_{\pm 0.0359}$ & $\underline{0.9719_{\pm 0.0008}}$ & $\textbf{0.9828}_{\pm 0.0005}$      \\
                            & $\Phi_{F1}$   $\uparrow$      & $0.5655_{\pm 0.0054}$ & $0.5624_{\pm 0.0044}$ & $0.9162_{\pm 0.0260}$ & $0.9124_{\pm 0.0463}$ & $\underline{0.9721_{\pm 0.0008}}$ & $\textbf{0.9828}_{\pm 0.0005}$      \\ 
\midrule
\multirow{3}{*}{Flower102}  & $\mathcal{A}$ $\uparrow$      & $0.6365_{\pm 0.0421}$ & $0.7462_{\pm 0.0258}$ & $0.8628_{\pm 0.0826}$ & $\underline{0.9025_{\pm 0.0394}}$ & $0.8829_{\pm 0.0215}$ & $\textbf{0.9098}_{\pm 0.0251}$      \\
                            & $\Phi_{P}$    $\uparrow$      & $0.6649_{\pm 0.0419}$ & $0.7992_{\pm 0.0350}$ & $\underline{0.9026_{\pm 0.0348}}$ & $0.9013_{\pm 0.0464}$ & $0.8734_{\pm 0.0063}$ & $\textbf{0.9175}_{\pm 0.0224}$      \\
                            & $\Phi_{R}$    $\uparrow$      & $0.6916_{\pm 0.0594}$ & $0.8209_{\pm 0.0408}$ & $\underline{0.9132_{\pm 0.0323}}$ & $0.9051_{\pm 0.0420}$ & $0.8977_{\pm 0.0143}$ & $\textbf{0.9289}_{\pm 0.0205}$      \\
                            & $\Phi_{F1}$   $\uparrow$      & $0.6421_{\pm 0.0435}$ & $0.7902_{\pm 0.0361}$ & $0.8872_{\pm 0.0485}$ & $0.8883_{\pm 0.0525}$ & $\underline{0.8696_{\pm 0.0135}}$ & $\textbf{0.9132}_{\pm 0.0253}$      \\ 
\midrule
\multirow{3}{*}{FGVC}       & $\mathcal{A}$ $\uparrow$      & $0.3369_{\pm 0.0182}$ & $0.3476_{\pm 0.0216}$ & $0.7682_{\pm 0.0193}$ & $0.7661_{\pm 0.0065}$ & $\underline{0.7841_{\pm 0.0089}}$ & $\textbf{0.8082}_{\pm 0.0199}$      \\
                            & $\Phi_{P}$    $\uparrow$      & $0.3512_{\pm 0.0225}$ & $0.3633_{\pm 0.0257}$ & $0.7324_{\pm 0.0511}$ & $0.7932_{\pm 0.0032}$ & $\underline{0.8007_{\pm 0.0034}}$ & $\textbf{0.8225}_{\pm 0.0119}$      \\
                            & $\Phi_{R}$    $\uparrow$      & $0.3499_{\pm 0.0128}$ & $0.3646_{\pm 0.0239}$ & $0.7387_{\pm 0.0256}$ & $0.7404_{\pm 0.0089}$ & $\underline{0.7657_{\pm 0.0131}}$ & $\textbf{0.8014}_{\pm 0.0292}$      \\
                            & $\Phi_{F1}$   $\uparrow$      & $0.3338_{\pm 0.0185}$ & $0.3480_{\pm 0.0228}$ & $0.7063_{\pm 0.0224}$ & $0.7328_{\pm 0.0093}$ & $\underline{0.7408_{\pm 0.0134}}$ & $\textbf{0.7842}_{\pm 0.0336}$      \\ 
\midrule
\multirow{3}{*}{StanfordCars}   & $\mathcal{A}$ $\uparrow$    & $0.2844_{\pm 0.0076}$ & $0.2842_{\pm 0.0123}$ & $\underline{0.9635_{\pm 0.0063}}$ & $0.9519_{\pm 0.0164}$ & $0.9590_{\pm 0.0025}$ & $\textbf{0.9721}_{\pm 0.0032}$      \\
                                & $\Phi_{P}$    $\uparrow$    & $0.3190_{\pm 0.0056}$ & $0.3125_{\pm 0.0048}$ & $0.9624_{\pm 0.0078}$ & $0.9596_{\pm 0.0125}$ & $\underline{0.9640_{\pm 0.0021}}$ & $\textbf{0.9751}_{\pm 0.0026}$      \\
                                & $\Phi_{R}$    $\uparrow$    & $0.2822_{\pm 0.0084}$ & $0.2823_{\pm 0.0124}$ & $\underline{0.9636_{\pm 0.0064}}$ & $0.9506_{\pm 0.0168}$ & $0.9598_{\pm 0.0024}$ & $\textbf{0.9716}_{\pm 0.0033}$      \\
                                & $\Phi_{F1}$   $\uparrow$    & $0.2861_{\pm 0.0090}$ & $0.2833_{\pm 0.0092}$ & $\underline{0.9619_{\pm 0.0076}}$ & $0.9505_{\pm 0.0172}$ & $0.9586_{\pm 0.0025}$ & $\textbf{0.9720}_{\pm 0.0032}$      \\ 
\midrule
\multirow{3}{*}{Average}    & $\mathcal{A}$ $\uparrow$      & $0.4947_{\pm 0.1394}$ & $0.5245_{\pm 0.1621}$ & $0.8948_{\pm 0.0746}$ & $0.8974_{\pm 0.0734}$ & $\underline{0.9003_{\pm 0.0618}}$ & $\textbf{0.9299}_{\pm 0.0634}$     \\
                            & $\Phi_{P}$    $\uparrow$      & $0.5339_{\pm 0.1433}$ & $0.5665_{\pm 0.1739}$ & $0.8907_{\pm 0.0873}$ & $\underline{0.8991_{\pm 0.0709}}$ & $0.8950_{\pm 0.0588}$ & $\textbf{0.9340}_{\pm 0.0571}$     \\
                            & $\Phi_{P}$    $\uparrow$      & $0.5310_{\pm 0.1591}$ & $0.5613_{\pm 0.1877}$ & $0.8947_{\pm 0.0809}$ & $0.8902_{\pm 0.0776}$ & $\underline{0.9091_{\pm 0.0730}}$ & $\textbf{0.9314}_{\pm 0.0646}$     \\

                            & $\Phi_{F1}$   $\uparrow$      & $0.5111_{\pm 0.1449}$ & $0.5452_{\pm 0.1767}$ & $0.8803_{\pm 0.0915}$ & $0.8794_{\pm 0.0820}$ & $\underline{0.8880_{\pm 0.0761}}$ & $\textbf{0.9247}_{\pm 0.0704}$     \\ 
\bottomrule
\end{tabular}
}
\caption{Open-vocabulary classification performance with different schemes. We report Accuracy $\mathcal{A}$, Precision $\Phi_{P}$, Recall $\Phi_{R}$ and F1 score $\Phi_{F1}$. \ours achieves the superior performance over all baseline methods.}
\label{tab:main}
\end{table*}

\subsection{Experimental Setup}
\paragraph{Dataset} 
We use 6 different image classification datasets in our experiments. They cover a wide range of classification challenges, which includes Caltech101\cite{fei2004learning} for generic objects classification; Food101\cite{bossard2014food}, Flowers102\cite{nilsback2008automated},  StanfordCars\cite{krause20133d} and FGVCAircraft\cite{maji2013fine} for fine-grained classification; UCF101 \cite{soomro2012ucf101} for action recognition. 

\paragraph{Baseline algorithms and models} We compare \ours against to two groups  of methods. The first group is federated learning with traditional neural networks: (1) FedAvg; (2)FedKA. {FedKA} is a state-of-the-art federated domain generalization method based on feature distribution matching. For both FedAvg and FedKA, we use a ResNet-18\cite{he2016deep} pre-trained on ImageNet\cite{deng2009imagenet}. The second group of baselines are methods that combine CLIP and FL: (1) PromptFL, a federated prompt tuning method; (2) TPG, a federated text-driven prompt generation method; (3) FedCLIP, a federated adapter-style finetuning method. For PromptFL, TPG, FedCLIP, as well as \ours, CLIP with configuration of ViT-L/14@336px is selected as the backbone model. For all methods, the aggregated global model is used for the evaluation on all different datasets.

\paragraph{Federated learning setup} To simulate the open-vocabulary setting, we split the classes of each dataset into two groups, one as training classes and the other as test classes. The data from training classes are available for local model training, whereas the images from test classes are only available during test time. Moreover, we consider a non-i.i.d. heterogeneous FL setting {as in \cite{qiu2023text}}. The training classes are disjointly distributed to different clients. That is, the classes of one client is mutually exclusive with the classes of any other clients. In a real-world application, it is usually hard for all clients to collect a huge amount of data. As such, we also consider a data-sparse setting, where all clients only have a few images per class for training as in \cite{qiu2023text}. The data is distributed over 10 clients, and there are 10 training images per class for all datasets (2 for validation). All samples of test classes are used for validation (20\%) and test (80\%). In robustness study, we modified the amount of training images per class. We repeat experiments for 5 times and report the mean and standard deviation in all tables. Further implementation details are in Appendix A.

\begin{table}[t]
\resizebox{\columnwidth}{!}{
\begin{tabular}{@{}lccccc@{}}
\toprule
\textbf{Dataset} & \textbf{Metrics}    & \textbf{\ours} & w/o A. A. & w/o M. P.   \\ 
\midrule
\multirow{4}{*}{Caltech101}
& $\mathcal{A}$ $\uparrow$      & $0.9936_{\pm 0.0010}$ & $0.9857_{\pm 0.0029}$ & $0.9332_{\pm 0.0197}$ \\
& $\Phi_{P}$    $\uparrow$      & $0.9848_{\pm 0.0030}$ & $0.9700_{\pm 0.0058}$ & $0.8898_{\pm 0.0219}$ \\
& $\Phi_{R}$    $\uparrow$      & $0.9908_{\pm 0.0014}$ & $0.9894_{\pm 0.0020}$ & $0.9784_{\pm 0.0042}$ \\
& $\Phi_{F1}$   $\uparrow$      & $0.9876_{\pm 0.0020}$ & $0.9790_{\pm 0.0038}$ & $0.9238_{\pm 0.0174}$ \\
\midrule
\multirow{4}{*}{UCF101}
& $\mathcal{A}$ $\uparrow$      & $0.9127_{\pm 0.0225}$ & $0.9073_{\pm 0.0352}$ & $0.8818_{\pm 0.0100}$ \\
& $\Phi_{P}$    $\uparrow$      & $0.9212_{\pm 0.0238}$ & $0.9105_{\pm 0.0374}$ & $0.8911_{\pm 0.0126}$ \\
& $\Phi_{R}$    $\uparrow$      & $0.9127_{\pm 0.0255}$ & $0.9073_{\pm 0.0352}$ & $0.8818_{\pm 0.0100}$ \\
& $\Phi_{F1}$   $\uparrow$      & $0.9086_{\pm 0.0298}$ & $0.9013_{\pm 0.0408}$ & $0.8702_{\pm 0.0127}$ \\
% \midrule
% \multirow{4}{*}{FGVC}
% & $\mathcal{A}$ $\uparrow$      & $0.8082_{\pm 0.0199}$ &  &  \\
% & $\Phi_{P}$    $\uparrow$      & $0.8225_{\pm 0.0119}$ &  &  \\
% & $\Phi_{R}$    $\uparrow$      & $0.8014_{\pm 0.0292}$ &  &  \\
% & $\Phi_{F1}$   $\uparrow$      & $0.7842_{\pm 0.0336}$ &  &  \\
\midrule
\multirow{4}{*}{Food101}    
& $\mathcal{A}$ $\uparrow$      & $0.9828_{\pm 0.0005}$ & $0.9827_{\pm 0.0006}$ & $0.9718_{\pm 0.0005}$ \\
& $\Phi_{P}$    $\uparrow$      & $0.9829_{\pm 0.0005}$ & $0.9828_{\pm 0.0006}$ & $0.9731_{\pm 0.0005}$ \\
& $\Phi_{R}$    $\uparrow$      & $0.9828_{\pm 0.0005}$ & $0.9827_{\pm 0.0006}$ & $0.9718_{\pm 0.0005}$ \\
& $\Phi_{F1}$   $\uparrow$      & $0.9828_{\pm 0.0005}$ & $0.9827_{\pm 0.0006}$ & $0.9720_{\pm 0.0005}$ \\ 
\midrule
\multirow{4}{*}{Flower102}  
& $\mathcal{A}$ $\uparrow$      & $0.9098_{\pm 0.0251}$ & $0.9003_{\pm 0.0340}$ & $0.8736_{\pm 0.0240}$    \\
& $\Phi_{P}$    $\uparrow$      & $0.9175_{\pm 0.0224}$ & $0.8886_{\pm 0.0353}$ & $0.8729_{\pm 0.0102}$    \\
& $\Phi_{R}$    $\uparrow$      & $0.9289_{\pm 0.0205}$ & $0.9123_{\pm 0.0319}$ & $0.8945_{\pm 0.0131}$    \\
& $\Phi_{F1}$   $\uparrow$      & $0.9132_{\pm 0.0253}$ & $0.8875_{\pm 0.0391}$ & $0.8684_{\pm 0.0131}$    \\ 
% \midrule
% \multirow{4}{*}{StanfordCars}   
% & $\mathcal{A}$ $\uparrow$    & $0.9721_{\pm 0.0032}$      \\
% & $\Phi_{P}$    $\uparrow$    & $0.9751_{\pm 0.0026}$      \\
% & $\Phi_{R}$    $\uparrow$    & $0.9716_{\pm 0.0033}$      \\
% & $\Phi_{F1}$   $\uparrow$    & $0.9720_{\pm 0.0032}$      \\                                 
\bottomrule
\end{tabular}
}
\caption{Ablation Study.}
\label{tab:ablation}
\end{table}

\subsection{Open-vocabulary Generalization}
We report the main results on open-vocabulary generalization for all baselines and datasets in Table~\ref{tab:main}. 
% In Table~\ref{tab:main}, we report the accuracy $\mathcal{A}$, Precision $\Phi_{P}$, Recall $\Phi_{R}$ and F1 score $\Phi_{F1}$ to demonstrate the model performance. 
The best results are highlighted in bold and the second-best results are highlighted with underlines. We observe: (1) Traditional FL methods could not address the open-vocabulary challenge. For example, {FedKA only achieves an averaged accuracy of 0.5245 over all datasets.} (2) \ours outperforms baselines on all datasets w.r.t. all metrics. For instance, on accuracy, \ours outperforms the best baseline by 3\% on average. (3) Across different datasets, \ours consistently demonstrates superior performance, while the baseline methods are sensitive to different datasets. For instance, PromptFL could achieve comparable accuracy of 0.9920 as \ours's 0.9936 on Caltech101. However, on UCF101, PromptFL only achieves 0.8582 accuracy, which is significantly lower than \ours with 0.9127. {We attribute such sensitivity to the unreliable generalization ability of the baselines, as they are not deliberately designed for open-vocabulary settings.} (4) Across different metrics, \ours consistently outperforms baselines, whereas the baselines are sensitive to the evaluation metrics. For instance, on Flower102, PromptFL achieves a high precision of 0.9026, but a low accuracy of 0.8628. Similarly, on the same dataset, TPG achieves a high accuracy of 0.9025, but a low F1 score.

\subsection{Ablation Study}
Next, we conduct an ablation study to understand the functionality of adaptive aggregation (A. A.) and multimodal prototyping (M. P.) in \ours. Due to space limit, we report the results on 4 datasets. The results are shown in Table~\ref{tab:ablation}. We observe that removing either module could cause a degradation of the model performance. {For instance, without adaptive aggregation, the accuracy of \ours on Caltech101 drops from 0.9936 to 0.9857. After removing multimodal prototyping, the accuracy on Caltech101 drops to 0.9332.}

\begin{figure}[t]
\centering
\includegraphics[trim=0.5cm 0.6cm 1.4cm 0cm, clip, width=\textwidth]{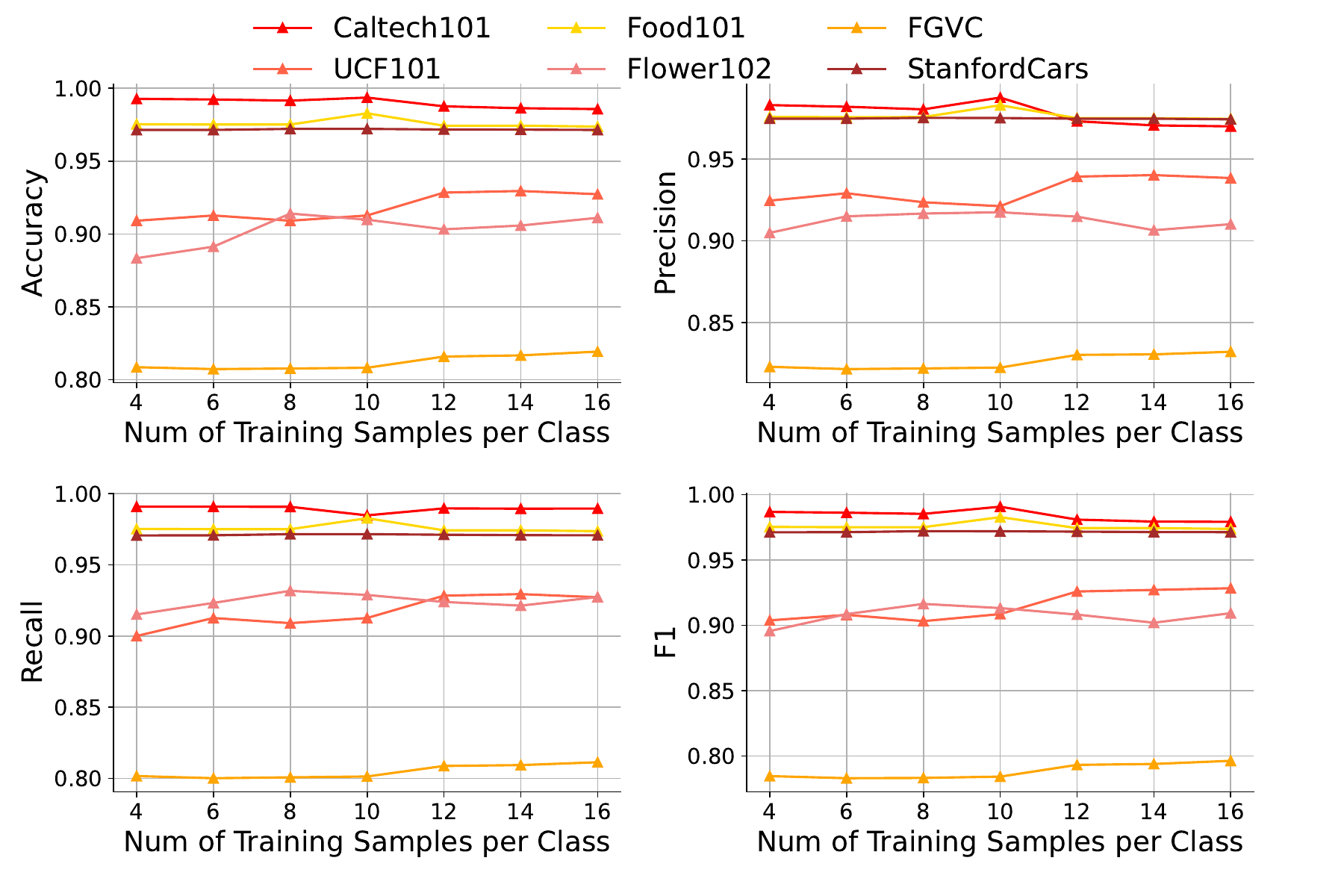}
\caption{Robustness study w.r.t. number of training samples.}
\label{fig:num_samples}
\end{figure}

\begin{figure}[t]
\centering
\includegraphics[trim=0.5cm 0.6cm 1.4cm 0cm, clip, width=\textwidth]{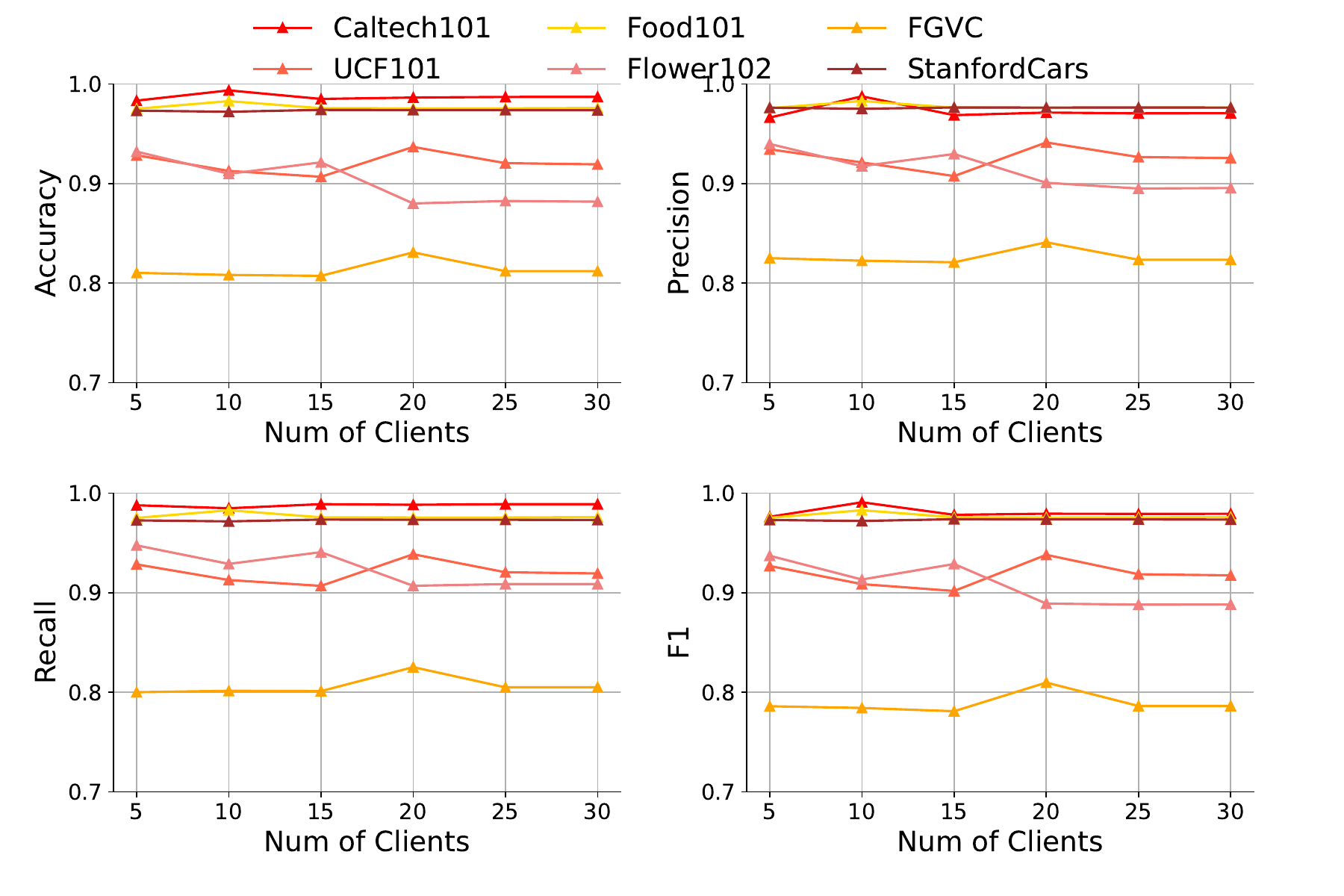}
\caption{Scalability study w.r.t. number of clients.}
\label{fig:num_clients}
\end{figure}

% \usepackage{booktabs}
% \usepackage{multirow}
\begin{table*}[t]
\small
\resizebox{0.8\textwidth}{!}{
\begin{tabular}{@{}lccccccc@{}}
\toprule
\textbf{Method}                     & FedAvg (NN) & FedKA (NN) & PromptFL & FedTPG & FedCLIP & \ours (ours) \\ 
\midrule
time/img (s)          & $0.0031$   & $0.0031$   & $0.0366$   & $0.0367$   & $0.0362$   & $0.0369$      \\
trainable params.       & $100.00\%$ & $100.00\%$ & $0.0140\%$ & $0.0549\%$ & $0.2745\%$ & $0.2783\%$      \\
\bottomrule
\end{tabular}
}
\caption{Averaged processing time per image (in seconds) and ratio of trainable parameters (in percentage).}
\label{tab:efficiency}
\vspace{-0.2cm}
\end{table*}

\subsection{Robustness Study}
In this section, we conduct a robustness study w.r.t. the number of training samples per class. {This is a key factor affecting the finetuning quality}. In particular, we change it from 2 to 16, and keep the number of clients as 10. The results are shown in Figure~\ref{fig:num_samples}. We observe that \ours is generally robust against the number of training samples. On Flower102 and FGVC, \ours is relatively more sensitive to the number of training samples. {This is because that different kinds of flowers and aircraft are more difficult to distinguish compared to food types and car makes.}

\subsection{Scalability Study}
\huimin{Then, we investigate the scalability of \ours w.r.t. the number of clients, as the number of clients in an FL application is a key factor affecting data heterogeneity and training stability. Specifically, we use the same test classes in our 10-client setting as the test classes, but re-distribute the training classes to different number of clients (from 5 to 30). The results are shown in Figure~\ref{fig:num_clients}. We observe that \ours is scalable and achieves consistent high performance as the number of clients increases.}

\subsection{Efficiency Evaluation}
\huimin{Finally, we compare the averaged processing time per image (in seconds) and the ratio of trainable parameters (in percentage) among different methods. We show that \ours is a light-weight and feasible solution for FL applications. From Table~\ref{tab:efficiency}, we observe CLIP-based methods need more time to process image than NN-based methods. Moreover, by comparing the processing time per image across different methods, our \ours is in general as fast as other CLIP-based baseline methods. In terms of memory saving, \ours can be considered as a parameter-efficient method, because only less than 0.3\% of the model parameters need to be trained, which is much fewer than FedTPG. However, we also acknowledge that \ours needs to train more parameters than FedCLIP. This is expected, because \ours uses the same adapter for the visual encoder as FedCLIP, but \ours also trains the extra client residuals.}

\section{Conclusion}
\label{sec:conclusion}
This work is the first to address the open-vocabulary challenge in FL applications. In particular, we present \ours, a novel open-vocabulary FL framework that is tailored for finetuning VLMs for FL applications. \ours provides an effective solution to make high-quality predictions for queries that involve novel unseen categories. {Extensive experimental results on various datasets demonstrate the effectiveness of our method.}

% We highlight the significance of this work: despite the promising integration of large foundation models with FL applications, existing literature largely overlooks the open-vocabulary challenge, which is of great practical meaning. 
\section{Limitations}
\label{sec:limitations}
One limitation of this work is that our method introduces extra hyperparameters. For different applications, one might need to finetune these hyperparameters, which brings extra computational cost. {As for the actually trainable modules, there is only a small two-layer network and light-weight perturbations.} Another limitation of this work is that our method does not take the inherent bias of the pre-trained VLM into account. However, it is known that the pre-trained foundation models usually have encoded the bias in the pre-training data (e.g., stereotypical data, racism and hate speech). Such bias could have negative ethical implications on downstream FL applications. Therefore, a future research direction is to develop a benign, fair, open-vocabulary FL framework. 
\section*{Ethics Statement}
Our work provides a data-efficient and privacy-aware solution to address the open-vocabulary problem in federated learning. Our method automatically generalizes to a new user and is capable of answering her/his queries that involve data from novel categories. {In terms of real-world applications, with \ours, the update frequency of the deployed FL model could be drastically reduced, and there is no need to collect huge amount of training data for novel classes. The above two advantages of \ours reduce the risk of collecting private user data.}

% This is beneficial in real-world applications, since keeping updating the system is infeasible due to the huge amount of potential users, and our method could rescue.

\section*{Acknowledgment}

This research is supported in part by the National Science Foundation under Grant No. IIS-2202481, CHE-2105032, IIS-2130263, CNS-2131622, CNS-2140999. The views and conclusions contained in this document are those of the authors and should not be interpreted as representing the official policies, either expressed or implied, of the U.S. Government. The U.S. Government is authorized to reproduce and distribute reprints for Government purposes notwithstanding any copyright notation here on.


% \clearpage

\bibliography{reference}

\appendix
\onecolumn
\section*{Appendix A: Implementation Details}
\paragraph{Hyperparameters} For \ours and all baseline methods that use CLIP, the learning rate is initialized as 1e-5. The learning rate for baseline methods that use ResNet-18 is 5e-4. The models are optimized via AdamW. The local training epoch is 2 and the global epoch is also 2. For all methods with key hyperparameters, we firstly performed grid search with the resolution of 0.1 until find the best performance. Based on that, we further reduce the search resolution to 0.01 until find best performance. In terms of the confidence threshold $\epsilon$, on Caltech101, UCF101, Flower102, we use $20\%$ of the maximum entropy given the distribution of the datasets on different clients. As for FGVC, Food101, we set $\epsilon$ equal to $3å0\%$ of maximum entropy. For StanfordCars, we used $10\%$. Our hardware is NVIDIA A40. 

\paragraph{Baseline Implementation} We use ImageNet pre-trained ResNet-18 as the backbone model for FedAvg and FedKA. Upon implementation, we modify and re-train the classification head of the pre-trained ResNet-18 to fit it into our classification problem. Moreover, when performing aggregation and inference, these classification heads are not used, because they can not provide predictions for unseen classes. Therefore, we only aggregate the feature extraction modules of the finetuned ResNet-18 to obtain the global model. As for inference, we use the aggregated feature extractor to produce adapted representations. Using extracted representations, we further perform K-means clustering and linear sum assignment, to map the representations onto the unseen test classes. K-means and linear sum assignment is implemented using the SciPy library.

\paragraph{Evaluation Metrics} In Table~\ref{tab:main}, we use the scikit-learn library to compute the macro-averaged F1. Due to class imbalance, it is likely that F1 score is lower than precision and recall at the same time.

\paragraph{Implementation of Multimodal Prototyping} Finally, when implementing multimodal prototyping, we do not save all the visual prototypes for the sake of efficiency. Instead, we only dynamically update and save the centroid of each visual prototype set. For each class, this could be done with following steps: 
\begin{itemize}
    \item At time step $n$, the centroids of all prototypes are computed;
    \item Save the centroids and the number of prototypes used for each class;
    \item At the next time step $n+1$, if there is a new prototype added to the prototype set of a specific class $c$, then the sum of previous prototypes of will be reproduced by $\sum_{q\in Q_c}q=\bar{q}_c \cdot |Q_c|$; 
    \item Update the new centroid of the visual prototype for class $c$: $\bar{q}_c = \frac{\sum_{Q_c}q + \frac{z'}{||z'||}}{|Q_c|+1}$.
\end{itemize}

\clearpage

\section{Appendix B: Overall Framework}
\begin{algorithm}[h]
\caption{\ours (Training)}
\label{alg:training}
\SetAlgoLined
\textbf{Input} CLIP image encoder $f_{I}$, CLIP text encoder $f_{T}$, adapter $f_{A}$, datasets of local clients $\mathcal{D}_1, \mathcal{D}_2, ..., \mathcal{D}_K$\;
\textbf{Hyperparameters} Learning rate;
Initialize the visual adapter $f_A$ \;
Clients download $f_{I}$, $f_{T}$ and $f_A$ \;
\For{k=1,2,...,K}{ 
    Receive trainable models: $f^{(k)}_{A} = f_{A}$ \;
    Initialize the client residual $\Delta^{(k)}$ \;
    \For{local epochs}{
        Compute normal visual representations: $z = f_I(x)$ \;
        Compute adapted visual representations: $z' = z + f_A(z)$\;
        Compute normal text representations: $t = f_T(\texttt{A photo of [class c]})$ \;
        Compute perturbed text representations: $t' = t + \alpha \delta$\;
        Compute CLIP adaptation loss $\mathcal{L}_{adap}$ with Equation~\ref{eq:adap_loss}\; 
        Update $f^{(k)}_A$ and $\Delta^{(k)}$ with gradient descent;
        }
    Obtain perturbed text representations $\mathcal{T}^{'(k)}$ by adding $\delta \in \Delta^{(k)}$ to $t$.
    }
\textbf{Output} Send $f^{(k)}_A$ and $\mathcal{T}^{'(k)}$ to the central server \;
\end{algorithm}

\begin{algorithm}[h]
\caption{\ours (Inference)}
\label{alg:inference}
\SetAlgoLined
\textbf{Input} CLIP image encoder $f_{I}$, CLIP text encoder $f_{T}$, adapter weights $f^{(1)}_{A}$, $f^{(2)}_{A}$,...$f^{(K)}_{A}$, perturbed client text representations $\mathcal{T}^{'(1)}$, $\mathcal{T}^{'(2)}$,...,$\mathcal{T}^{'(K)}$, test data $\mathcal{D}_{test}$, test prompts $\mathcal{T}_{test}$\;
\textbf{Hyperparameters} Confidence threhold $\epsilon$;
Compute the expected similarity between the test user and clients using Equation~\ref{eq:simi}\;
Obtain $f_A$ by aggregating the adapter weights using Equation~\ref{eq:fed_avg}\; 
Initialize the visual prototypes as empty sets \;
\For{$x$ $\in \mathcal{D}_{test}$}{ 
    Compute the centroids for the visual prototypes with Equation~\ref{eq:centriods}\;
    Compute the prediction with Equation~\ref{eq:prototype_inference}\;
    Update the corresponding visual prototype set using the original pseudo prediction and Equation~\ref{eq:prototype}\;
    }
\textbf{Output} Predictions for $\mathcal{D}_{test}$ \;
\end{algorithm}

% \clearpage
% \section{Appendix C: Robustness Study w.r.t. Number of clients.}
% \input{tables/robustness_study}

\end{document}